# DEVELOPMENT OF THE CHATGPT, GENERATIVE ARTIFICIAL INTELLIGENCE AND NATURAL LARGE LANGUAGE MODELS FOR ACCOUNTABLE REPORTING AND USE (CANGARU) GUIDELINES

Giovanni E. Cacciamani[1,*], Michael B. Eppler,[1] Conner Ganjavi,[1] Asli Pekan[1], Brett Biedermann[1], Gary S. Collins[2], Inderbir S. Gill[1]

[1]University of Southern California, Los Angeles, CA, USA
[2] UK EQUATOR Centre, Centre for Statistics in Medicine, Nuffield Department of Orthopaedics, Rheumatology and Musculoskeletal Sciences, University of Oxford, Oxford, UK.

*University of Southern California IRB Approval*
IRB#: UP-23-00306

**\*Principal Investigator and Corresponding author**
Giovanni Cacciamani MSc, MD, FEBU
Keck School of Medicine
University of Southern California
Los Angeles, CA, USA
Giovanni.cacciamani@med.usc.edu

**ABSTRACT**

The swift progress and ubiquitous adoption of Generative AI (GAI), Generative Pre-trained Transformers (GPTs), and large language models (LLMs) like ChatGPT, have spurred queries about their ethical application, use, and disclosure in scholarly research and scientific productions. A few publishers and journals have recently created their own sets of rules; however, the absence of a unified approach may lead to a 'Babel Tower Effect,' potentially resulting in confusion rather than desired standardization. In response to this, we present the ChatGPT, Generative Artificial Intelligence, and Natural Large Language Models for Accountable Reporting and Use Guidelines (CANGARU) initiative, with the aim of fostering a cross-disciplinary consensus on the ethical use, disclosure, and proper reporting of GAI/GPT/LLM technologies in academia.

The present protocol consists of four distinct parts: a) an ongoing systematic review of GAI/GPT/LLM applications to understand the linked ideas, findings, and reporting standards in scholarly research, and to formulate guidelines for its use and disclosure, b) a bibliometric analysis of existing author guidelines in journals that mention GAI/GPT/LLM, with the goal of evaluating existing guidelines, analyzing the disparity in their recommendations, and identifying common rules that can be brought into the Delphi consensus process, c) a Delphi survey to establish agreement on the items for the guidelines, ensuring principled GAI/GPT/LLM use, disclosure, and reporting in academia, and d) the subsequent development and dissemination of the finalized guidelines and their supplementary explanation and elaboration documents.

**INTRODUCTION**

The rapid advancement and widespread uptake of Generative Pre-trained Transformer (GPTs) and more generally large language models (LLMs), such as ChatGPT and others, including academia, have brought forth questions regarding their responsible and ethical use, application and disclosure academic research and scientific output [1, 2]. Generative-AI (GAI), using Large Language Models like OpenAI's ChatGPT launched on November 30, 2022 [3], has swiftly gained traction. Within two months of its launch, ChatGPT amassed 100 million monthly users, becoming the fastest adopted technology in history [4], and eventually boosting the development and release of new AI tools. This success spurred similar offerings from tech giants and open-source platforms alike, such as Google's Bard and Microsoft's Bing Chat [5, 6], MedPalm2 [7, 8] and many others [9].

Over 650 research articles and editorials have since explored GAI tools as ChatGPT and other powered by LLMs over the past 6 months, lauding its capabilities in enhancing academic writing and research processes, such as improving grammar[10], translating text[11], generating novel research ideas [10], and synthesizing large data[12], coding and images creation. Despite these advantages, caution is advised as GAI/GPTs/LLMs cannot be held legally responsible for its outputs and there exists potential (amongst others) for inaccuracies, perpetuating existing bias, and plagiarism [12, 13].

Over the past few months, publishers and journals have independently started proposing guidelines to protect research integrity and provide guidance to researchers, editors, reviewers, and authors [1, 14-17]. However, uniform guidelines and methods for GAI/GPTs/LLM reporting in the scholarly community are essential to prevent the possible 'Babel Tower Effect' arising when multiple parties create their own bespoke guidance or regulations, potentially leading to disarray, instead of standardization. Although any document proposing guidance comes from good intentions, it could be unproductive or even hazardous if not thoughtfully devised. It is therefore timely to develop a comprehensive cross-specialty recommendations for the ethical and responsible use of GAI/GPTs/LLMs in academic research and scientific writing [18]. Uniting a diverse group of professionals – scientists, clinicians, researchers, statisticians, computer scientists, engineers, methodologists, regulatory committees, and journal editors – we aim to create a collaborative community focused on developing inclusive, globally-relevant recommendations[19]. Following established and standardized methodology for developing consensus-based reporting guidelines[18], we aim to produce a robust and inclusive framework.

    Herein we propose the ChatGPT, Generative Artificial Intelligence and Natural Large Language Models for Accountable Reporting and Use Guidelines (CANGARU) initiative to enable cross-discipline consensus on accountable use, disclosure, and guidance for reporting of GAI/GPTs/LLMs usage in academia [15, 19]. Working collaboratively with individuals, academic and publishing regulatory organizations to develop a uniform set of recommendations will effectively mitigate the potential confusion or conflicting guidance that could occur when multiple groups independently work on the same task. In turn, this will ensure any recommendations are coherent, comprehensive, and universally applicable, promoting responsible usage of this powerful technology.

**AIM AND OBJETIVES**

The present study has been approved by the University of Southern California Institutional Review Board (IRB#: UP-23-00306). The overall aim of the CANGARU initiative is to establish guidance on commonly shared, cross-discipline best practices for using GAI/GPTs/LLMs in academia as follows:

    a. **The 'DON'T' Criteria List:** The 'DON'T' Criteria List will be a comprehensive guideline that aims to ensure ethical and proper use of GAI/GPTs/LLMs in

academic research. It will cover each step of the academic process, starting from (but not limited too) generating research hypotheses, study conduct and data analysis (including coding), image creation (see later comment), interpretation of findings, and drafting and reviewing, refining and editing the manuscript. This list will serve as a valuable resource to researchers, providing guidance on what to avoid throughout the various stages of the research process. By identifying and highlighting potential pitfalls and ethical concerns, the 'DON'T' Criteria List will help researchers navigate the use of GAI/GPTs/LLMs responsibly and with integrity.

b. **Disclosure Criteria List:** The Disclosure Criteria List will provide guidance for researchers to transparently disclose their use of GAI/GPTs/LLMs in academic research, improving transparency, accountability, and enabling better assessment of research findings. It emphasizes the importance of what and how to disclose, fostering responsibility and addressing potential risk and limitations associated with this technology.

c. **Reporting Criteria List:** The reporting criteria list will provide a checklist of recommendation to ensure the complete and transparent reporting of GAI/GPTs/LLMs when they are used as interventions in scientific studies. It will consider all the important aspects that should be reported in the scientific manuscript in order to enhance transparency, improve reproducibility, standardize reporting practices, reduce misinterpretation, support peer review and editorial processes, and facilitate research synthesis and knowledge translation.

Specific core objectives are as follows:

i. Conduct a living systematic review on GAI/GPTs/LLMs to 1) gain a thorough understanding of the ideas and observations related to the use of GAI/GPTs/LLMs in academic research and scientific writing and 2) collect and generate ideas for the proposed guidelines regarding its usage as well as how it should be disclosed in publications, 3) collect common reporting criteria.

ii. Conduct a bibliometric analysis of existing journal instruction to authors that mention GAI/GPTs/LLMs tools to 1) evaluate how publishers and journals have already released their own guidelines for using, disclosing, and reporting GAI/GPTs/LLMs , 2) to assess the heterogeneity in the recommendations in terms of what is allowed, what should be disclosed, and what should be reported, and 3) to identify common recommendations to be used for contributing to and informing the statements to be rated during the Delphi consensus process (step iii).

iii. Conduct a Delphi survey to identify and reach consensus on the items for the above-mentioned guidance list (a b and c, as above mentioned) to ensure the accountable use, disclosure, and reporting of GAI/GPTs/LLMs technology in academia.

iv. Creation and dissemination of the guidance and accompanying Explanation and Elaboration documents.

Note: These guidelines pertain to a snapshot of the current state of this technology and will be updated as validation studies address the concerns or issues mentioned in this document.

**METHODS AND DESIGN**

The CANGARU Guidelines development will follow the EQUATOR (Enhancing the QUAlity and Transparency Of health Research) Network's framework for developing reporting guidelines[20]. The CANGARU project was registered as 'Guideline under development' on the EQUATOR Network website in March 2023. The project will be overseen by a Core Team (CT). The CT will meet regularly and be responsible for project operations and the ultimate advancement of the project. It is comprised of a group of international experts that will collaborate to provide guidance and project strategy.

The CT comprise diverse group of professionals, including scientists, clinicians, researchers, statisticians, computer scientists, engineers, methodologists, and journal editors, from top-ranked journals according to SCImago.org, as well as representatives from academic publishers, COPE (Committee on Publication Ethics), EASE (European Association of Science Editors), and WAME (World Association of Medical Editors), EQUATOR Network and STM (AI Ethics), to establish a community collaboration that will create inclusive reporting guidelines for AI with a global perspective and to enhance scholarly and cross-disciplinary inclusiveness.

The project will be carried out in seven phases (*figure 1*): as follow:

    (1) Identification of key areas to be addressed through the Delphi Consensus

        (1.1) Living systematic literature review on GAI/GPTs/LLMs,

        (1.2) Bibliometric analysis of the existing guidelines on GAI/GPTs/LLMs in academia,

        (1.3) Collection of Core team members perspectives]

    (2) Identification of CANGARU Guidelines statements for each list,

    (3) Delphi Experts Panel survey,

    (4) Consensus Meeting,

(5) Piloting,

(6) Checklist and statement explanation and elaboration document, and

(7) Dissemination.

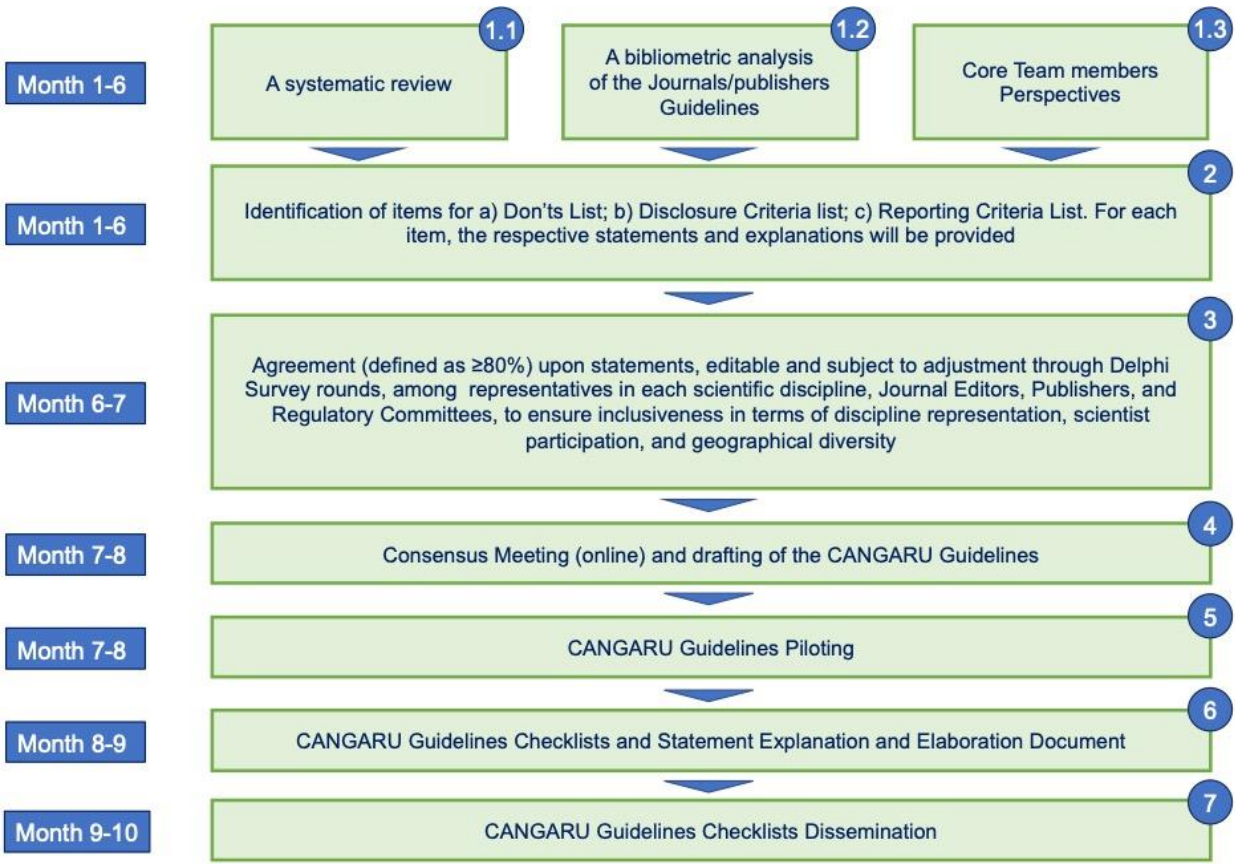

Figure 1. CANGARU development flowchart and timeline

**Phase 1 - Identification of candidate issues that will form the Delphi survey.**

Phase 1 involves the identification of the key areas for the three lists as reported in the aim and objective paragraph using three different sources, as reported below:

*1.1 Living Systematic Review*

We have started a living systematic review, with the aim to capture all publications related to GAI/GPTs/LLMs prospectively at 6, 12 and 24 months after the release of ChatGPT (December 5th, 2022). Two Independent reviewers will perform the abstract and full-text screen, and each article will be reviewed for data extraction by at least 2 reviewers. Article title, first author, published year and journal will be automatically exported into excel. Journal discipline is based on listed disciplines on SCImago. Articles will be reviewed for the following information:

a) Tasks, benefits, and risks of GAI/GPTs/LLMs will be identified and extracted and assigned into categories.

b) Studies will be identified that explicitly disclosed the use of GAI/GPTs/LLMs in an acknowledgment/disclosure section.

c) For studies that used GAI/GPTs/LLMs as an intervention, reporting criteria will be collected.

The reporting of the living systematic review will follow PRISMA guidelines[21] and is registered with PROSPERO (registration ID: CRD42023406875).

*The objectives of this prospective systematic review are to 1) gain a thorough understanding of the issues and observations related to the use of* GAI/GPTs/LLMs *in scientific research and 2) generate a list of candidate issues for potential inclusion in the guidelines regarding its usage, as well as how it should be disclosed in publications, 3) collect common reporting criteria.*

### Phase 1.2. Bibliometric Analysis of the Existing Journal Instructions to Authors on GAI/GPTs/LLMs *in scientific publishing*

We have started a bibliometric analysis of the existing reporting and disclosure guidelines (in journal instructions to authors) on the use of GAI/ GPTs/LLMs powered tools to evaluate the current status and heterogeneity of recommendations on this matter. Two reviewers will screen independently the author's guidelines websites of the top 100 scientific publishers and the top 100 journals, according to the list of top-100 Scientific publishers [22] and SCImago Journal & Country Rank (SJR) portal (www.scimagojr.com) respectively. The collection of the information will be performed in 24 hours. The official websites for each Publisher and journal will be manually searched for author instructions. Data regarding the presence of GAI/GPTs/LLMs use/disclosure/reporting guidelines will be independently collected into a database under the supervision of a senior investigator. After discussion, any dispute will be resolved by consensus and adjudication by the senior investigator. The data collected will include any general or specific recommendations or references for the use, disclosure, or reporting of GAI/GPTs/LLMs. Publishers' and journals' guidelines referring specifically to other third-party regulatory agencies will also be included.

*The objectives of this bibliometric analysis are 1) to evaluate how publishers and journals have already released their own guidance for using, disclosing, and reporting GAI/GPTs/LLMs in scientific manuscripts, 2) to assess the heterogeneity in the recommendations in terms of what is allowed, what should be disclosed, and what should be reported, and 3) to identify common recommendations and concerns, which will included in the Delphi consensus process and inform the development of the guidance statements.*

**Phase 1.3. Core-Team Blind Members' Perspectives**
Each member of the core team not involved in the systematic review, or the bibliometric analysis will be asked to provide criteria that would be suitable for the inclusion in any of the three lists as follows. For each topic presented, members of the core team will be asked to contextualize them, articulating the underlying reasons.

    a. The 'DON'T' Criteria List, as defined above.

    b. Disclosure Criteria List, as defined above.

    c.    Reporting Criteria List, as defined above.

*The objective of this phase is to gather feedback from publishers, editors, guideline developers, and regulatory representatives regarding the potential impact of GAI/GPTs/LLMs on their respective academic fields, and to suggest strategies for promoting the responsible use of these technologies.*

**Phase 2. Identification of Potential Key Items to Be Included in the Delphi Survey**

The three sources mentioned in phases 1.1, 1.2 and 1.3 will be combined and duplicates candidate recommendations removed, resulting in three sets of statements informed by a systematic review of the literature, a bibliometric analysis of existing recommendations, and the perspectives of the CT members.

    During this phase, the recommendations and issues identified in stages 1.1, 1.2, and 1.3 are synthesized to form an extensive list of candidate items corresponding to each established criteria list. Subsequently, in phase 3, these items undergo a rating process designed to generate a degree of consensus. This consensus is then carried forward to phase 4 for further action and analysis.

*The objective of this phase is to generate a comprehensive set of items for each list (a, b and c) to be included in the Delphi Consensus Questionnaire.*

**Phase 3. Delphi Survey**

We will conduct a Delphi survey involving a broad global and cross-discipline stakeholder group. This survey will consist of up to three rounds, the aim of which is to reach consensus on what elements should be incorporated in each of the aforementioned criteria lists. The Delphi technique is an approach for facilitating expert dialogue concerning complex issues to achieve a collective agreement. The Delphi method serves as an effective communication tool for both individual group members and the collective group[23]. It circumvents the potential for interpersonal conflicts while promoting independent thinking among stakeholders[24].

*The objective of this phase is to achieve consensus on a selection of crucial elements for each of the three lists (a, b, c as previously mentioned). This step aims to ensure the sample's global coverage and its representation across various academic disciplines.*

*Design*

The Delphi survey will entail multiple stages, in which panelists will individually and anonymously assess whether to agree, disagree, or suggest modifications to the suggested key items for each one of the three lists (a,b and c as reported above). This process will be executed for up to three rounds. After each round, participants will receive aggregated feedback from the preceding round to aid in harmonizing personal views and establishing group agreement. Items that receive strong agreement (≥80% endorsement) will proceed to the consensus meeting (Phase 4). Borderline items will be considered by the Core Team as to whether to take them forward to the consensus meeting. Participants will anonymously complete each round of the Delphi survey using the Welphi platform ([www.welphi.com](www.welphi.com)). Following each Delphi round, the survey results will be summarized utilizing the Welphi software.

*Selection of potential key items for Delphi survey*

See Phase 2.

*Selection of Delphi Survey participants*

It is beneficial to include participants with a range of experience and expertise[25]. Our goal, by seeking a broad array of expert perspectives, is to involve a diverse group of stakeholders to participate in the Delphi survey. This approach is designed to ensure that the guidance will have geographically diverse applicability across all disciplines and for every type of scientific study.

We plan to extend invitations to participants from various roles and settings globally, including those from industry and academia. Given the absence of a clear definition for expertise, particularly in emerging fields such as GAI/GPTs/LLMs quantifying expertise becomes a challenging task. To address this, we have devised a strategy to engage with unique groups of stakeholders who might be affected by these guidelines.

To ensure comprehensive representation across diverse perspectives and broad range of stakeholders, we will adopt a purposeful oversampling strategy. This strategy ensures that we include voices from various fields, geographies, and demographic categories, such as gender and ethnicity. Through these measures, we aim to curate a diverse and inclusive participant pool for our Delphi study, thereby enhancing the reliability and validity of our findings.

Our selection of participants will target global academic associations and institutions, as well as developers who have published works on GAI/GPTs/LLMs. We will also seek insights from AI-guidelines developers and experts specializing in the ethical aspects of AI. Additionally, we aim to involve researchers who have published articles in the top 5 journals across each of the 27 major thematic categories, as identified by Scimago.org, in the past 12 months.

The diverse stakeholder group will encompass academic professionals – scientists, clinicians, researchers, statisticians, computer scientists, engineers, methodologists, regulatory committees, publishers, and journal editors. Once stakeholders with special interests are identified, reached out to, and engaged, we will employ the snowball sampling method [26].

Consistent with the Delphi surveys used in developing other consensus-based reporting guidelines, [27-33], our aim is to include a minimum of 200 participants worldwide. CT members can suggest participants and/or identify international associations to ensure global inclusiveness. We will acquire participants' informed consent through a digital consent form, and they will retain the right to withdraw at any time. Those who decide to withdraw from the survey will not be sent any additional invitations. To preserve confidentiality, participants will not be privy to the identities of others in the Delphi panellists or be aware of the specific responses given by any panel member.

### *Delphi Rounds*

Participants will evaluate the applicability and quality of key items for each list (a,b and c as reported above) for incorporation into the reporting guidelines by means of a 1 to 5 Likert scale. A 5-point Likert scale has been chosen to gauge agreement on each key item [30, 34]. The numerical score represents the following:

1: Strong agreement

2: Mild agreement

3: Indecision

4: Mild disagreement

5: Strong disagreement

In addition to the Likert scale, each question will include a free-text area for participants to suggest ways to refine the proposed definition or propose extra elements. Participants will also be given the choice of 'unable/unwilling to answer' for any of the prompts.

At the end of the first round, the CT will compile the results. The second Delphi round will be circulated to participants who completed the first one. Upon receiving the second round Delphi, participants will be provided with the responses from the initial round along with corresponding relevant free-text suggestions. Participants will be given a period of 1 week to conclude the second round. Items that received strong agreement (≥80%) in the first round will not be re-voted but will be made available for context. Participants will get a chance to vote on items that were modified based on feedback from the first round. We anticipate that two rounds will suffice, however, there may be an additional 3rd round. Participants will be given a period of 1 week to conclude each round. The CT will examine the results from all Delphi rounds. All consensus-reaching criteria will be showcased at the consensus meeting, coupled with data on the agreement level. Moreover, elements that failed to reach consensus and pertinent suggestions gathered during the Delphi rounds might be discussed at the discretion of the CT, during the consensus meeting.

**Phase 4. Consensus Meeting**

The main objective of the consensus gathering is to finalize decisions regarding which items to incorporate into the CANGARU Guidelines. In addition, there should be an option to refine and improve definitions that gained consensus in the previous Delphi survey round and to reconsider items that did not reach consensus.

The CT's responsibility will be to ensure a diverse representation from all stakeholders. Once the participants are chosen, they will receive an email invitation, with details on the rules to achieve consensus, objectives, and agenda of the meeting. The meeting will be structured to allow for in-depth discussions and break-out sessions to ensure the most effective and comprehensive definitions are collectively agreed upon.

The attendees will comprise members of the CT and international stakeholders who have completed the Delphi Survey rounds. Participants who have completed both rounds of the Delphi survey will be given the chance to indicate their interest in attending the consensus meeting.

The phrasing of each set of key item for each list (a,b and c ad reported above) will be composed and ratified by the CT. The initial drafts of the 3 lists will be disseminated among the consensus meeting participants to validate that they accurately reflect the group's consensus.

*The objective of this phase is to finalize the definitions of the key items that will be included in the guidelines. This will be achieved through a consensus meeting with representatives from all stakeholders. The meeting will also provide an opportunity to refine and improve the definitions that have already been agreed upon.*

**Phase 5. CANGARU Guidelines Piloting**

Once the CANGARU Guideline have been finalized at the consensus meeting, the CT will draft and edit the document intended for broad dissemination for the piloting phase. Invitees to this phase will encompass members of the CT, those who took part in the Delphi survey and consensus meeting, and other researchers who weren't directly involved in the development of the CANGARU Guidelines. At this stage, participants will be requested to identify any potential weaknesses, exclusions, or ambiguous points in the CANGARU Guidelines through their practical application.

*The objective of this phase is to evaluate the practicality and receptiveness of the guidelines when applied in a real-world scenario.*

**Phase 6. CANGARU Guidelines Explanation and Elaboration document**

The CT will be responsible for leading the development of CANGARU Guideline Statements and the corresponding Explanation and Elaboration documents. The CT will have the discretion to make updates (such as removing, merging, splitting, or adding items) to the CANGARU Guidelines checklists during the development of the updated Statements, as deemed necessary based on the pilot testing (phase 5) and feed this back to the consensus meeting participants for sign off. The Explanation and Elaboration documents will be published, which will provide an explanation for including the three lists (a,b and c as mentioned above) and provide examples of reporting best practices. The explanation and elaboration papers will present the reasoning behind the proposed items in each and offer illustrative examples. Drafts of these papers will be shared with all CT members for their feedback and revision before final publication [30, 32].

*The purpose of this phase is to assess and refine the documents to ensure they are ready for global dissemination and endorsement.*

**Phase 7. CANGARU Guidelines Dissemination**

As soon as the checklist publication becomes publicly available, we will initiate a robust campaign to extensively distribute the CANGARU Guidelines. Our campaign will focus on disseminating the guidelines publication through social media platforms and at conferences. The CANGARU Guidelines will be published in conjunction with explanatory videos and will also be featured on the EQUATOR website ([www.equator-network.org](www.equator-network.org)). We will target publishers and journals to endorse these guidelines in an effort to reduce the confusion of differing standards. To bolster accessibility and foster inclusivity, we have planned for verified translations in multiple languages of the 3 check lists.

*This phase is crucial for maximizing awareness of these guidelines and ensuring their broad applicability.*

**CONCLUSION**

The proposed CANGARU Guidelines will comprise important recommendations that data scientists should consider for ethical accountability and future reproducibility. They will provide recommendations on what to avoid, what and how to disclose, and how to ensure proper reporting of GAI/GPTs/LLMs in papers. The guidelines aim to enable academics, authors, editors, reviewers, publishers, and readers to critically follow the relevant etiquette when using GAI/GPTs/LLMs.

The resultant CANGUARU Guidelines will:

1) Serve as a tool for training researchers on the ethical and accountable use of GAI/GPTs/LLMs in academic research and scientific writing.

2) Establish a framework for assessing publications for publishers, editors, authors, and reviewers.

3) Assist end-users, including scientific audiences and policymakers, in better evaluating the validity and applicability of scientific papers for their decision-making process.

*reporting adverse events during surgical procedures and evaluating their impact on the postoperative course.* European Urology Focus, 2022.